# Multimodal Prompt Learning with Dual-Granularity Contrastive Alignment for Preoperative Prediction of Lung Adenocarcinoma Invasiveness Using 3D CT

Qing Huang[a,b#], Xiaotong Tan[a,b#], Lintong Yao[c#], Wufeng Xue[a,b], Ruobing Huang[a,b], Dong Ni[a,b], Hao Wu[c*], Tian Xiao[d*], Jun Cheng[a,b*]

[a]National-Regional Key Technology Engineering Laboratory for Medical Ultrasound, Guangdong Key Laboratory for Biomedical Measurements and Ultrasound Imaging, School of Biomedical Engineering, Thoracic Surgery Department of the First Affiliated Hospital, Shenzhen University Medical School, Shenzhen University, Shenzhen, China

[b]Marshall Laboratory of Biomedical Engineering, Shenzhen University, Shenzhen, China

[c]Department of Thoracic Surgery, Shenzhen Second People's Hospital, The First Affiliated Hospital of Shenzhen University, Shenzhen 518035, Guangdong Province, China

[d]Guangdong Key Laboratory of Genome Instability and Human Disease Prevention, Department of Biochemistry and Cell Biology, School of Basic Medical Sciences, Shenzhen University Medical School, Shenzhen University, Shenzhen 518055, China

[#]Co-first authors

[*]Co-corresponding authors

# Abstract

Accurate preoperative prediction of lung adenocarcinoma invasiveness is important for surgical planning and individualized treatment. However, overlapping CT appearances make it challenging to distinguish invasive adenocarcinoma (IAD) from Pre-IAD lesions. We propose a multimodal prompt-learning framework with dual-granularity image-text alignment for 3D CT-based IAD classification. A 3D ResNet-CBAM encoder extracts global and spatial visual features. A frozen RadBERT backbone is adapted with separate learnable prompts for report-level and keyword-level text encoding. These prompts are initialized using semantic representations derived from real-world clinical reports. Global contrastive alignment connects whole-volume CT representations with reports generated from clinician-annotated radiological attributes to provide overall semantic supervision. A multiple instance learning-based local alignment module associates spatial CT features with vascular-focused radiological keywords, to capture local imaging characteristics potentially associated with tumor invasion. The framework jointly optimizes classification and alignment objectives, while inference requires only CT images. Experiments included 1,593 pathologically confirmed nodules. On the test set, the proposed method achieved an accuracy of 87.92%, F1-score of 87.78%, and AUC of 0.9499. Its AUC significantly exceeded that of the image-only Res-CBAM baseline with the same visual backbone (0.9297; DeLong test, $p = 0.0078$). Ablation analyses supported the complementary contributions of global and local alignment and the benefit of report-based prompt initialization. Qualitative visualizations showed activation within nodules and adjacent vascular structures. These findings suggest that dual-granularity radiological supervision can enhance CT-based invasiveness prediction while preserving a CT-only inference workflow.

## 1. Introduction

Lung cancer remains the leading cause of cancer-related mortality worldwide [1], and lung adenocarcinoma is its most common histological subtype [2]. According to the classification of the International Association for the Study of Lung Cancer, atypical adenomatous hyperplasia (AAH), adenocarcinoma in situ (AIS), minimally invasive adenocarcinoma (MIA), and invasive adenocarcinoma (IAD) represent a spectrum of lung adenocarcinoma progression [3]. For clinical assessment of invasiveness, AAH, AIS, and MIA are commonly grouped as Pre-IAD, whereas IAD represents the invasive subtype [4]. Patients with Pre-IAD generally have favorable postoperative outcomes and may be candidates for limited surgical resection, such as wedge resection or segmentectomy [5]. In contrast, IAD is associated with a higher risk of recurrence and often requires more extensive surgical management [6, 7]. Distinguishing these groups before surgery can therefore inform surgical planning and prognostic assessment. However, substantial overlap in CT manifestations between Pre-IAD and IAD makes reliable assessment of tumor invasiveness challenging in clinical practice [8].

To address this challenge, radiomics studies have used quantitative CT features to predict the invasiveness of pulmonary nodules [9–11]. Although these methods have demonstrated promising results, they typically rely on handcrafted imaging features and multi-stage analysis pipelines, which may limit their ability to fully exploit the rich information in CT images. Deep learning methods have therefore been increasingly explored for invasiveness prediction. For example, Pan et al. proposed an adjudication strategy–based framework that combines multiple binary and ternary classifiers for invasiveness prediction and demonstrated robust performance on external datasets [12]. Recently, vision-language learning have created new opportunities for medical image analysis. By aligning visual representations with language-based clinical knowledge, vision-language models can capture richer semantic information than conventional image-only approaches. For example, Lei et al. proposed CLIP-Lung to predict lung nodule malignancy, which uses a channel-wise conditional prompt and multi-view contrastive learning guided by nodule textual attributes [13].

Despite this progress, many deep learning methods for IAD prediction are trained primarily with pathological labels and rely on image-derived features, with limited explicit use of radiological knowledge. This may limit their ability to represent subtle imaging findings associated with early invasion and can make the learned features difficult to interpret clinically. Meanwhile, existing methods often focus on global nodule representations and lack explicit modeling of fine-grained local imaging signs related to invasiveness [14–16]. The rich semantic information in radiological text also remains underused as a source of supervision for visual representation learning. Although vision-language learning has shown promise in medical image analysis, its application to 3D CT-based IAD classification remains underexplored. Effectively aligning 3D visual representations with radiological text remains challenging, as global image–text alignment may not adequately capture fine-grained spatial correspondences between local CT features and specific radiological signs.

To address these issues, this paper proposes a multimodal prompt learning framework with dual-granularity image-text alignment for IAD classification using 3D CT. Built upon a frozen RadBERT [17] backbone with learnable prompts initialized from real-world clinical reports, the framework constructs two separate prompt-based text encoders to achieve global report-level contrastive alignment and local keyword-level alignment with 3D CT features, respectively. For local alignment, a multiple instance learning (MIL) pooling strategy is introduced to explicitly associate spatial CT features with vascular-related radiological keyword embeddings, encouraging the model to capture local features potentially associated with tumor invasion. With this design, radiological knowledge provides supervision for visual representation learning during training, while the inference pipeline requires only 3D CT volumes. This may facilitate integration of the framework into CT-based assessment workflows.

The main contributions of this work are summarized as follows:

- We propose a dual-granularity image–text alignment framework for preoperative IAD classification using 3D CT. The framework combines global report-level and local keyword-level supervision to guide visual representation learning while retaining CT-only inference.
- We introduce an MIL-based local alignment module that associates spatial CT features with radiological keyword embeddings under nodule-level attribute supervision. This design enables fine-grained semantic alignment without requiring spatial annotations of individual radiological signs.
- We develop a knowledge-informed prompt-learning strategy that initializes learnable prefixes using semantic representations derived from real-world clinical radiology reports. Separate prefixes adapt a frozen RadBERT backbone for global report-level and local keyword-level encoding.

## 2. Related work

### 2.1 CT-Based Prediction of Lung Adenocarcinoma Invasiveness

CT-based prediction of lung adenocarcinoma invasiveness has been extensively investigated to support preoperative risk stratification. Radiomics-based approaches typically combine quantitative imaging features with clinical and radiological characteristics. For example, Liu et al. developed a radiomics-based nomogram integrating CT radiomic signatures and clinical-radiological features [18]. Deep learning approaches have also been explored to capture more complex imaging patterns. Zhou et al. developed an ensemble multi-view 3D CNN to differentiate Pre-IAD from IAD and further stratify IAD by pathological grade, achieving performance comparable to or exceeding that of experienced radiologists [4]. Wan et al. developed a ternary classification model integrating nodule size, mean CT attenuation, and morphological features to differentiate preinvasive lesions, MIA and IAD, achieving a C-index of 0.92 [19]. Zhang et al. further combined 3D CT representations with radiologist-annotated imaging signs using cross-attention and adaptive modality gating for multiclass invasiveness stratification [20] .

These studies have demonstrated the value of deep learning and complementary radiological information for predicting lung adenocarcinoma invasiveness. However, most existing approaches primarily exploit CT-derived features or explicitly annotated imaging signs, leaving the semantic information contained in radiology reports underexplored. This motivates the incorporation of report-level semantics and specific radiological attributes into visual feature learning for improving Pre-IAD/IAD classification.

## 2.2 Medical Vision–Language Alignment

Vision-language alignment has emerged as an effective approach to incorporating textual semantics into medical image analysis. Li et al. used large language models to extract diagnostic labels from radiology reports and employed these labels to enhance visual representation learning for 3D chest CT [21]. Moving beyond global semantic supervision, Yan et al. explored joint global and local alignment for few-shot medical image classification and out-of-distribution detection [22]. MedAligner focused on fine-grained alignment through word-region correspondence and diagnostic term filtering [23]. Within a unified pretraining framework, Zhang et al. developed MedUnifier, which combines image-text contrastive learning, image-text matching, and text-guided image generation [24], Alignment strategies have also been developed to incorporate relationships across imaging views and time points. For multi-view mammography, Du et al. introduced GLAM, which integrates geometry-guided local alignment with global vision-language and cross-view visual contrastive learning [25]. Lian et al. proposed ALTA, a parameter-efficient adaptation framework that incorporates temporal and multi-view radiographs to improve medical vision-language alignment for retrieval and zero-shot classification [26].

Together, these studies demonstrate the potential of textual supervision at different semantic and spatial levels. Their application to 3D CT-based IAD classification, particularly the joint alignment of global nodule representations with report semantics and spatial CT features with radiological keywords, remains underexplored.

## 2.3 Prompt Learning for Medical Vision–Language Models

Prompt learning provides a parameter-efficient approach to adapting pretrained models to downstream tasks. Methods such as CoOp [27] and CoCoOp [28] learn task-specific textual prompts while keeping the pretrained model parameters fixed. In medical image analysis, Zhao et al. proposed RC-TPL, which constructs dual textual prompts from radiology reports and medical knowledge for explainable zero-shot X-ray diagnosis and lesion localization [29]. Bie et al. developed XCoOp, which uses large language models to generate clinical concept-driven prompts and aligns them with learnable prompts and medical image representations at multiple granularities [30]. Similarly, Fang et al. introduced ViP, a visual symptom-guided prompt learning framework that uses LLM-generated disease descriptions to guide learnable context and merge prompts for medical image classification [31]. Koleilat et al. proposed BiomedCoOp, which combines LLM-generated prompt ensembles, semantic consistency constraints, and knowledge distillation for few-shot biomedical image classification [32].

These approaches illustrate how clinically informed prompts can support the adaptation of pretrained models to medical image analysis tasks. Motivated by these studies, our framework uses separate learnable prompts initialized using semantic representations derived from real-world clinical reports to adapt a frozen RadBERT backbone for report-level and keyword-level alignment with 3D CT features.

## 3. Materials and Methods

### 3.1 Dataset

This retrospective study included patients with surgically resected pulmonary nodules who underwent surgical treatment at The Second People's Hospital of Shenzhen between January 2018 and June 2024. The inclusion criteria were as follows: (1) availability of preoperative thin-slice chest CT scans in DICOM format within 3 months before surgery; (2) availability of corresponding radiology reports; and (3) postoperative pathological confirmation of lung adenocarcinoma with an identifiable pathological subtype. Each nodule was treated as an independent sample for analysis.

Figure 1 shows the sequential screening procedure. The initial cohort comprised 1,896 patients with 2,013 nodules. Fifty-eight nodules from 52 patients were excluded because of non-primary lung adenocarcinoma, a previous history of lung cancer, inadequate CT quality, or severe motion artifacts. A further 362 nodules were excluded because of diameter greater than 30 mm (n = 78), multiple nodules in the same lobe that could not be matched to pathological findings (n = 42), incomplete pathological records (n = 219), or special histological subtypes (n = 23). The final dataset comprised 1,593 nodules, including 863 Pre-IAD and 730 IAD nodules.

The dataset was randomly divided into training, validation, and test sets at an approximate ratio of 7:1.5:1.5, comprising 1114, 239, and 240 nodules, respectively. To prevent data leakage across sets, all nodules from the same patient were assigned to the same set. The training set was used to train the models, the validation set for hyperparameter tuning and model selection, and the test set for final performance evaluation. Demographic and pathological characteristics are summarized in Table 1.

## 3.2 Overview of the Proposed Framework

The proposed framework uses radiological text to guide visual representation learning for preoperative classification of Pre-IAD and IAD (Figure 2). It combines global report-level alignment with local keyword-level alignment: reports provide an overall description of the nodule, while radiological attributes provide more specific supervision for spatial CT features.

Each training sample comprises a 3D CT volume, clinician-annotated radiological attributes, and a report generated from these attributes. A 3D visual encoder extracts global and spatial representations from the CT volume. Two text encoders built on a frozen RadBERT backbone use separate learnable prefixes to encode the report into a global representation and the radiological keywords into keyword-level embeddings. The learnable prefixes adapt the text representations to the task while keeping the pretrained backbone frozen.

Global contrastive alignment matches the global CT representation with the corresponding report representation. Local alignment uses MIL to aggregate similarities between spatial CT features and keyword embeddings, supervised by binary labels indicating attribute presence or absence.

Training combines four objectives: image classification, text classification, global contrastive alignment, and local MIL-based alignment. Text provides supervision during training, whereas inference requires only CT images, with predictions obtained from the visual encoder and its classification head.

### 3.3 Visual Encoder

We employ a 3D ResNet [33] backbone augmented with the Convolutional Block Attention Module (CBAM) [34] to encode the input CT volume V into a feature map:

$$F = f_v(V) \in \mathbb{R}^{C \times Z' \times H' \times W'}, \tag{1}$$

where $f_v(\cdot)$ denotes the visual encoder, C is the number of channels, and Z', H', and W' are the spatial dimensions.

We derive both global and spatial representations from $F$ for the two alignment branches. Global average pooling over the spatial dimensions produces a C-dimensional global representation:

$$f_g^v = GAP(F). \tag{2}$$

This representation summarizes the CT features for image classification and global image–text alignment.

For local alignment, a learnable linear transformation projects the C-dimensional feature vector at each spatial location into a d-dimensional visual token $f_i$. The spatial dimensions are then flattened to form a sequence of tokens:

$$f_{flat} = \{f_i\}_{\{i=1\}}^{\{N\}}, \qquad N = Z' \times H' \times W'. \tag{3}$$

Each token corresponds to one spatial location in the feature map, allowing local CT features to be compared with radiological keyword embeddings in the local alignment module.

## 3.4 Radiological Keyword Construction

We defined 12 radiological keywords representing CT features associated with lung adenocarcinoma invasiveness, based on clinician annotations and radiology reports (Figure 3). These features covered nodule density, morphology, vascular findings, and pleural involvement. For the local alignment experiments, the keywords were organized into four overlapping subgroups: morphology-focused, density-focused, pleural-focused, and vascular-focused. Group names indicate their primary imaging focus rather than an exclusive feature category. The composition of each group is shown in Figure 3.

Different keyword subgroups and the all-keyword group were compared as inputs to the local alignment module. The vascular-focused group was selected for the final configuration. For global alignment, a large language model (DeepSeek-V3.2 [35] ) generated a structured radiology report (Figure 2) for each nodule from its annotated attributes using the complete 12-keyword set.

### 3.5 Prompt-based Text Encoder

Two types of radiological text were used in this study. Real-world clinical radiology reports were used to derive semantic representations for prompt initialization, whereas structured reports constructed from clinician-annotated radiological attributes were used for report-level alignment during model training.

Given the constructed radiology report $T$ and the clinician-annotated keyword set $K = \{k_m\}_{m=1}^{M}$, the prompt-based text encoders are employed to extract semantic representations at both global and local levels.

The encoders are built upon RadBERT, whose parameters are frozen during training. To adapt the model to the target task, a prefix-tuning strategy is adopted by introducing learnable prompt tokens into each Transformer layer. For the $l$-th Transformer layer, the learnable prefixes are prepended to the key and value sequences during self-attention computation:

$$Attention(Q, K, V) = Softmax\left(\frac{\left(Q\left[K_p{}^{(l)};\ K\right]^T\right)}{\sqrt{d}}\right)\left[V_p{}^{(l)};\ V\right]. \tag{4}$$

To provide clinically meaningful knowledge injection for prompt learning, the prompt tokens are initialized using representations derived from real-world radiology reports. Reports from the training set are first encoded using the pretrained RadBERT model, and the resulting embeddings are clustered to capture common semantic patterns in clinical descriptions. The cluster centroids are then used to initialize the learnable prompt tokens, which are subsequently optimized during training.

To capture multi-level semantics, two text encoders with shared backbone but different learnable prefixes are constructed. The report encoder maps the generated report $T$ to a global text representation:

$$f_g^t = f_t^{(g)}(T), \tag{5}$$

which summarizes the overall semantic content of the report for global image–text alignment and text classification. The keyword encoder maps the radiological keyword set K to keyword-level embeddings:

$$f_k^{(m)} = f_t^{(k)}(k_m), m = 1, \dots, M, \tag{6}$$

where $M$ is the number of keywords used for local alignment and $k_m$ represents the m-th radiological attribute.

### 3.6 Training Objectives

We formulate a text-guided training objective that combines classification with global and local alignment. Clinically meaningful textual information provides supervision for visual representation learning at two complementary levels.

At the global level, report-level semantic representations provide holistic contextual supervision, encouraging the visual encoder to capture overall diagnostic patterns. At the local level, keyword-level embeddings derived from radiological attributes act as fine-grained supervision signals, guiding spatial visual features to focus on localized imaging signs associated with tumor invasion. Together, these objectives are designed to incorporate both overall report semantics and specific radiological signs into visual representation learning.

### 3.6.1 Global Alignment

We align the global visual representation $f_g^v$ with the global textual representation $f_g^t$ to encourage high-level semantic consistency between CT images and radiology reports. Within each mini-batch, the global contrastive learning loss is defined as:

$$\mathcal{L}_{global} = -\frac{1}{B}\sum_{i=1}^{B} \log \frac{\exp\left(f_g^{v(i)} \cdot f_g^{t(i)}/\tau\right)}{\sum_{j=1}^{B} \exp\left(f_g^{v(i)} \cdot f_g^{t(j)}/\tau\right)}, \tag{7}$$

where $B$ denotes the batch size and $\tau$ is a temperature parameter. For each CT volume, its corresponding report serves as the positive match, while reports from other samples in the mini-batch serve as negative matches. Unlike a symmetric image–text contrastive objective, this loss uses an image-to-text matching direction. It encourages the visual representations to be more similar to their corresponding report representations than to unmatched reports, thereby providing report-level semantic supervision for visual feature learning.

### 3.6.2 Local Alignment

Complementary to global alignment, local alignment associates spatial CT features with radiological keywords to provide supervision for localized imaging signs.

We compute the similarity between each spatial visual token $f_i$ and each keyword embedding $f_k^{(m)}$:

$$s_{i,m} = f_i \cdot f_k^{(m)}. \tag{8}$$

This produces a similarity matrix in which each column corresponds to a keyword and represents its responses across all spatial locations.

To aggregate these spatial responses, we treat the spatial visual tokens $\{f_i\}_{i=1}^{N}$ as a bag of instances and adopt a MIL pooling strategy. For each keyword, the highest similarity is selected:

$$\tilde{s}_m = \max\ \{\boldsymbol{s_{i,m}} | i \in \{1, \dots, N\}\} / \tau, \tag{9}$$

where $N$ is the number of spatial visual tokens, $\tau$ is a temperature parameter. The pooled response represents the strongest match between the keyword and the spatial features.

The keyword embeddings represent clinically annotated radiological attributes, providing explicit semantic targets for local alignment. Given the binary label $y_m$ indicating whether the $m$-th attribute is present, the local alignment loss is defined as:

$$\mathcal{L}_{local} = -\frac{1}{M}\sum_{m=1}^{M}[y_m \log \sigma(\tilde{s}_m) + (1 - y_m)\log\big(1 - \sigma(\tilde{s}_m)\big)], \tag{10}$$

where $M$ is the number of keywords and $\sigma(\cdot)$ denotes the sigmoid function. This loss encourages high pooled responses for present attributes and low responses for absent attributes, guiding spatial visual features to encode information relevant to the annotated radiological signs.

### 3.6.3 Classification Objectives

In addition to alignment, both global visual and textual representations are supervised using the pathological class label $y$. The image classification loss is defined as:

$$\mathcal{L}_{img} = CE\left(f_g^v, y\right), \tag{11}$$

and the text classification loss is defined as:

$$\mathcal{L}_{text} = CE\left(f_g^t, y\right), \tag{12}$$

where $f_g^v$ and $f_g^t$ denote the global visual and textual representations, respectively, and CE denotes cross-entropy loss.

The image classification objective directly supervises invasiveness prediction, while the text classification objective encourages the report representations to retain information relevant to the same task. Joint optimization combines class-label supervision with semantic alignment between the visual and textual representations.

### 3.6.4 Overall Objective

The overall training objective is a weighted sum of the four losses:

$$\mathcal{L} = \lambda_1 \mathcal{L}_{img} + \lambda_2 \mathcal{L}_{global} + \lambda_3 \mathcal{L}_{local} + \lambda_4 \mathcal{L}_{text}, \tag{13}$$

where the coefficients control the relative contributions of the four objectives. Together, these objectives supervise classification while incorporating global report semantics and local radiological attributes into visual representation learning.

## 3.7 Implementation Details

The cropped 3D CT volumes containing the nodules were resized to 32×32×32 voxels. The loss weights were set to $\lambda_1 = 0.5$, $\lambda_2 = 0.2$, $\lambda_3 = 1.0$, and $\lambda_4 = 0.8$. The model was trained using AdamW with learning rates of $1\times10^{-4}$ for the visual encoder and $1\times10^{-3}$ for the prompt parameters, and a batch size of 64. Model training was performed on an NVIDIA GeForce RTX 3090 GPU for 50 epochs. Hyperparameters were determined through grid search based on validation performance, and the final model was selected according to the lowest validation loss

Classification performance was evaluated using accuracy, precision, recall, F1-score, and the area under the receiver operating characteristic curve (AUC), with IAD defined as the positive class. Differences in AUC between competing methods were assessed using the DeLong's test, with $p < 0.05$ considered statistically significant.

# 4. Results

## 4.1 Overall Classification Performance

The proposed framework was compared with representative 3D CNN-based architectures (3D-ResNet [36], DenseNet121 [37], SEResNet50 [38], and Res-CBAM), a 3D Transformer-based model (ViT [39]), and two deep learning methods for lung nodule analysis (Deep-Lung [40] and NAS-Lung [41]). On the test set, it yielded the highest accuracy (0.8792), precision (0.8799), F1-score (0.8778), and AUC (0.9499) among the evaluated methods, with a recall of 0.8766 (Table 2).

Compared with the conventional 3D-ResNet baseline, the proposed framework improved accuracy from 0.8542 to 0.8792 and AUC from 0.9269 to 0.9499. It also outperformed Res-CBAM, which used the same visual backbone without textual supervision, across all five metrics. The AUC increased from 0.9297 to 0.9499, and this difference was statistically significant (DeLong's test, p=0.0078). These results suggest that text-guided dual-granularity alignment contributed to the performance improvement beyond architectural enhancements alone.

NAS-Lung achieved the highest recall (0.9154), but its precision was lower than that of the proposed framework (0.8095 versus 0.8799). The proposed framework showed a more balanced combination of precision and recall, yielding the highest F1-score among the evaluated methods.

## 4.2 Module Ablation Analysis

We conducted ablation experiments to assess the contribution of individual components by removing an alignment objective or model component. The results are summarized in Table 3.

Removing global alignment reduced AUC from 0.9499 to 0.9401, whereas removing local alignment reduced it to 0.9294. Both alignment objectives therefore contributed to discrimination, with a larger AUC decrease observed after removing local alignment. These findings support the complementary roles of global report-level and local keyword-level supervision.

We also examined the contribution of report-based prompt initialization. Replacing this initialization with random initialization in the local or global text branch reduced AUC to 0.9472 and 0.9455, respectively. Randomly initializing the prompts in both branches reduced AUC to 0.9394. These results suggest that report-based initialization benefited both branches. However, random initialization in the local branch yielded the same accuracy as the complete framework (0.8792) and a slightly higher F1-score (0.8782 versus 0.8778).

In addition, removing the learnable prompts and using the frozen text encoders directly reduced AUC to 0.9345, suggesting that learnable prompts play an important role in adapting the frozen language model to the target task. Removing CBAM reduced AUC to 0.9400, while omitting the text classification loss reduced it to 0.9464. These findings suggest that attention-based visual feature enhancement and text classification supervision contributed to the observed performance. The complete framework achieved the highest AUC among all evaluated configurations.

### 4.3 Influence of Radiological Keyword Groups

We compared four keyword subgroups—morphology-focused, density-focused, pleural-focused, and vascular-focused—and the complete keyword set as inputs to the local alignment module. The results are summarized in Table 4.

The vascular-focused subgroup achieved the highest performance across all five metrics: (accuracy 0.8792, precision 0.8799, recall 0.8766, F1-score 0.8778, and AUC 0.9499). Using the complete keyword set yielded a similar AUC of 0.9498, with an accuracy of 0.8750. The morphology-focused, density-focused, and pleural-focused subgroups yielded AUCs of 0.9462, 0.9457, and 0.9451, respectively.

These results suggest that the vascular-focused subgroup provided sufficient keyword-level supervision for local alignment, with performance comparable to that obtained using the complete keyword set. Adding the remaining attributes did not yield further improvement in these experiments.

### 4.4 Influence of Prompt Length

We evaluated prompt lengths of 2, 4, 8, and 16 to examine the effect of the number of learnable prefix tokens on classification performance (Table 5). A prompt length of 4 achieved the highest performance across all five metrics, with an accuracy of 0.8792 and AUC of 0.9499. The corresponding AUCs for lengths of 2, 8, and 16 were 0.9342, 0.9453, and 0.9326, respectively. These results suggest that a relatively short prefix was sufficient for task adaptation, while increasing its length beyond 4 did not improve performance and introduced redundant parameters.

### 4.5 Comparison of Different Text Encoders

We further compared five pretrained biomedical language models for text encoding. As shown in Table 6, the proposed framework based on RadBERT achieved the highest performance across all five metrics, with an accuracy of 0.8792 and AUC of 0.9499. The corresponding AUCs were 0.9401 for BioBERT, 0.9341 for BioClinicalBERT, 0.9330 for BlueBERT, and 0.9326 for PubMedBERT. These results suggest that radiology-specific semantic representations may better support image–text alignment within the proposed framework.

### 4.6 Visual Interpretability Analysis

Representative Grad-CAM [46] visualizations comparing the image-only Res-CBAM baseline with the proposed text-guided framework are shown in Fig. 4. In the displayed cases, the proposed framework showed more extensive activation within the nodules and surrounding perinodular regions.

Res-CBAM showed relatively focal activation within portions of the nodules in several examples. In contrast, the proposed framework exhibited broader activation across central and peripheral nodule regions, with responses also extending to adjacent vascular structures. The activation maps showed broader coverage of heterogeneous nodule components, particularly in subsolid nodules.

These qualitative observations suggest that text-guided training may help the visual encoder incorporate information from both the nodules and their surrounding structures that are potentially relevant to tumor invasiveness.

## 5. Discussion

Accurate preoperative discrimination between Pre-IAD and IAD is important for treatment planning and prognostic assessment, but overlapping CT appearances make this task challenging. In this study, we developed a text-guided representation learning framework that uses radiological reports and keywords to supervise CT feature learning. The proposed framework achieved an AUC of 0.9499, compared with 0.9297 for the image-only Res-CBAM baseline using the same visual backbone (DeLong’s test, $p = 0.0078$). These findings support the value of incorporating radiological semantic supervision into visual representation learning for lung adenocarcinoma invasiveness prediction.

The ablation results support the complementary contributions of global report-level and local keyword-level alignment. Removing either objective reduced AUC, with a larger decrease observed after removing local alignment. This suggests that local keyword-level supervision made a greater contribution to discrimination within the evaluated framework. Report-level alignment encourages consistency between global visual features and overall report semantics, whereas local alignment links spatial visual features to specific radiological keywords. Such explicit supervision may provide more focused guidance for learning imaging patterns associated with invasiveness.

The keyword-group experiments further suggest that a focused subset of radiological features can provide useful local supervision. The vascular-focused group achieved an AUC of 0.9499, closely matching the performance obtained using all keywords (0.9498). These results suggest that a compact, clinically informed keyword set may provide sufficient guidance for local alignment without requiring the full set of annotated features.

An important characteristic of the proposed framework is that radiological text is used as supervision during training, while prediction requires only CT images. By incorporating textual semantic information into visual representation learning, the framework retains a CT-only inference workflow without requiring radiological reports or keyword annotations for new cases. This design reduces the input requirements at inference and may facilitate integration into image-based assessment workflows.

Several aspects warrant further investigation. First, although this study included a relatively large dataset of 1,593 pathologically confirmed nodules, it was conducted at single center. Future studies involving additional centers may further support the generalizability of the proposed framework across different clinical settings. Second, radiological reports generated from predefined imaging attributes provided consistent semantic supervision for model training. Incorporating routine clinical reports in future studies would allow evaluation of the framework across more diverse reporting styles. Third, the visualization analyses offered insights into the model's localization of diagnostically relevant regions. Further exploration of the relationship between these regions and pathological findings may provide additional understanding of the model's predictions.

## Acknowledgements

This work was supported by Shenzhen Medical Research Fund (No. D2602009), Guangdong Basic and Applied Basic Research Foundation (Nos. 2025A1515011821 and 2026A1515012488), Shenzhen Science and Technology Program (No. JCYJ20240813143302004), National Natural Science Foundation of China (Nos. 12326619 and 62171290), Science and Technology Planning Project of Guangdong Province (No. 2023A0505020002), and Frontier Technology Development Program of Jiangsu Province (No. BF2024078). The funding sources have no involvement in the study.

## Conflict of interest

The authors have no relevant conflicts of interest to disclose.

## Tables

**Table 1.** Demographic and pathological characteristics of the training, validation, and test sets.

| Characteristics | Training Set (n = 1114) | Validation Set (n = 239) | Test Set (n = 240) |
|---|---|---|---|
| Age (years) | | | |
| Median | 54 | 53 | 54 |
| Range | 15–85 | 23–85 | 18–83 |
| Sex, n (%) | | | |
| Female | 732 (65.7%) | 156 (65.3%) | 168 (70.0%) |
| Male | 382 (34.3%) | 83 (34.7%) | 72 (30.0%) |
| Pathological category, n (%) | | | |
| Pre-IAD | 603 (54.1%) | 130 (54.4%) | 130 (54.2%) |
| IAD | 511 (45.9%) | 109 (45.6%) | 110 (45.8%) |

Note: Sample sizes (n) indicate the number of nodules. Pre-IAD includes atypical adenomatous hyperplasia, adenocarcinoma in situ, and minimally invasive adenocarcinoma; IAD, invasive adenocarcinoma.

**Table 2.** Performance comparison of different methods for preoperative prediction of lung adenocarcinoma invasiveness.

| Model | Accuracy | Precision | Recall | F1-score | AUC |
|---|---|---|---|---|---|
| Res-CBAM | 0.8583 | 0.8602 | 0.8545 | 0.8564 | 0.9297 |
| ResNet | 0.8542 | 0.8578 | 0.8493 | 0.8517 | 0.9269 |
| Deep-Lung | 0.8375 | 0.8421 | 0.8318 | 0.8344 | 0.9327 |
| NAS-Lung | 0.8375 | 0.8095 | **0.9154** | 0.8592 | 0.9247 |
| DenseNet121 | 0.8500 | 0.8493 | 0.8483 | 0.8487 | 0.9225 |
| SEResNet50 | 0.8333 | 0.8337 | 0.8301 | 0.8314 | 0.9105 |
| ViT | 0.8542 | 0.8529 | 0.8542 | 0.8534 | 0.9268 |
| Ours | **0.8792** | **0.8799** | 0.8766 | **0.8778** | **0.9499** |

**Table 3.** Ablation analysis of different components in the proposed framework.

| Setting | Accuracy | Precision | Recall | F1 | AUC |
|---|---|---|---|---|---|
| w/o global alignment | 0.8667 | 0.8753 | 0.8601 | 0.8636 | 0.9401 |
| w/o local alignment | 0.8583 | 0.8571 | 0.8580 | 0.8575 | 0.9294 |
| w/o knowledge-informed prompt initialization in local alignment | 0.8792 | 0.8785 | **0.8780** | **0.8782** | 0.9472 |
| w/o knowledge-informed prompt initialization in global alignment | 0.8667 | 0.8733 | 0.8608 | 0.8639 | 0.9455 |
| w/o knowledge-informed prompt initialization | 0.8667 | 0.8657 | 0.8657 | 0.8657 | 0.9394 |
| w/o prompt tuning | 0.8458 | 0.8454 | 0.8437 | 0.8444 | 0.9345 |
| w/o CBAM | 0.8542 | 0.8680 | 0.8458 | 0.8498 | 0.9400 |
| w/o text classification loss | 0.8708 | 0.8701 | 0.8696 | 0.8698 | 0.9464 |
| Ours | **0.8792** | **0.8799** | 0.8766 | 0.8778 | **0.9499** |

**Table 4.** Classification performance using different keyword groups for local alignment.

| Keyword Group | Accuracy | Precision | Recall | F1 | AUC |
|---|---|---|---|---|---|
| Vascular-focused group | **0.8792** | **0.8799** | **0.8766** | **0.8778** | **0.9499** |
| Pleural-focused group | 0.8583 | 0.8646 | 0.8524 | 0.8554 | 0.9451 |
| Morphology-focused group | 0.8667 | 0.8733 | 0.8608 | 0.8639 | 0.9462 |
| Density-focused group | 0.8625 | 0.8639 | 0.8591 | 0.8607 | 0.9457 |
| Complete keyword set | 0.8750 | 0.8752 | 0.8727 | 0.8737 | 0.9498 |

**Table 5.** Effect of prompt length on classification performance.

| Prompt Length | Accuracy | Precision | Recall | F1 | AUC |
|---|---|---|---|---|---|
| 2 | 0.8542 | 0.8545 | 0.8514 | 0.8526 | 0.9342 |
| 4（ours） | **0.8792** | **0.8799** | **0.8766** | **0.8778** | **0.9499** |
| 8 | 0.8625 | 0.8639 | 0.8591 | 0.8607 | 0.9453 |
| 16 | 0.8458 | 0.8480 | 0.8416 | 0.8435 | 0.9326 |

**Table 6.** Classification performance of the proposed framework using different text encoders.

| Text Encoder | Accuracy | Precision | Recall | F1 | AUC |
|---|---|---|---|---|---|
| BioBERT-base-cased-v1.2[42] | 0.8667 | 0.8753 | 0.8601 | 0.8636 | 0.9401 |
| PubMedBERT[43] | 0.8375 | 0.8372 | 0.8395 | 0.8372 | 0.9326 |
| BioClinicalBERT[44] | 0.8667 | 0.8700 | 0.8622 | 0.8646 | 0.9341 |
| BlueBERT[45] | 0.8583 | 0.8646 | 0.8524 | 0.8554 | 0.9330 |
| RadBERT (ours) | **0.8792** | **0.8799** | **0.8766** | **0.8778** | **0.9499** |

# Figures

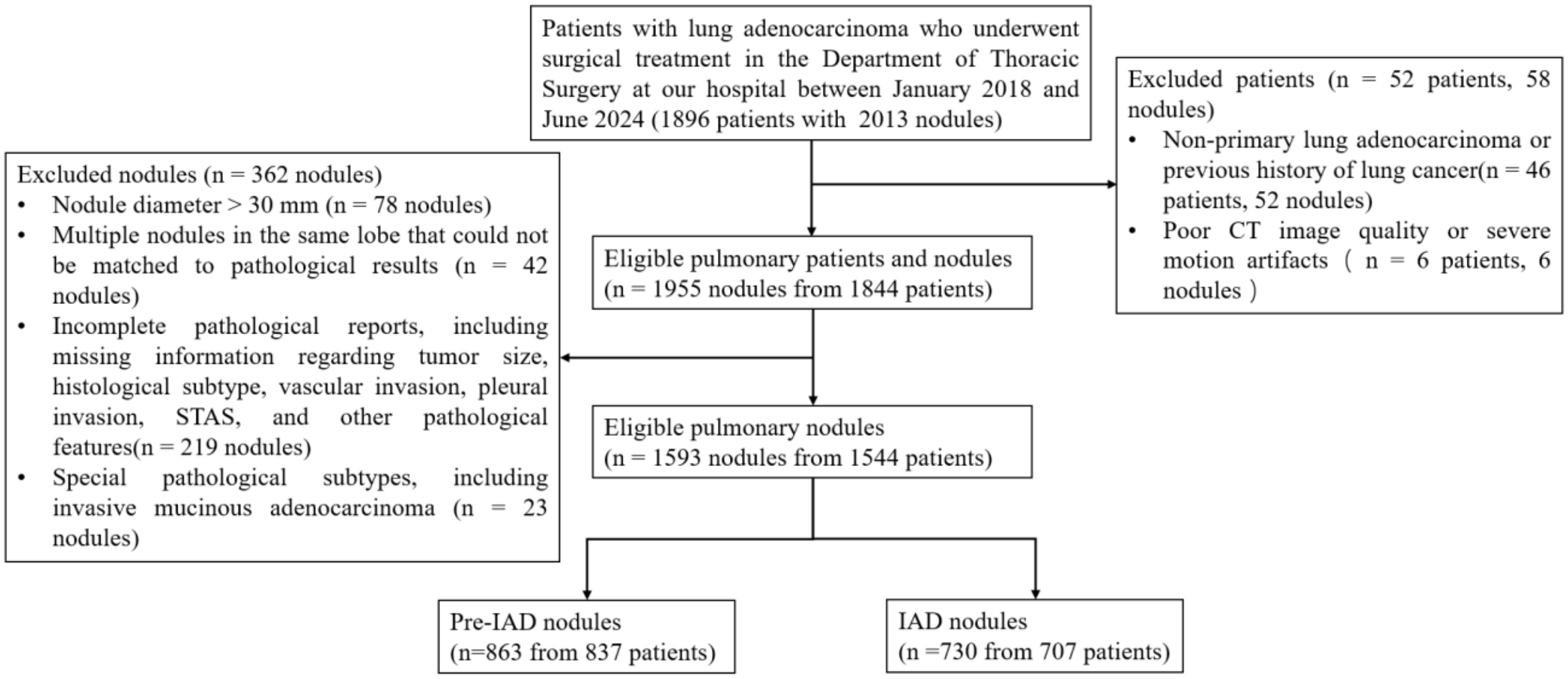

**Figure 1.** Flow diagram of cohort construction.

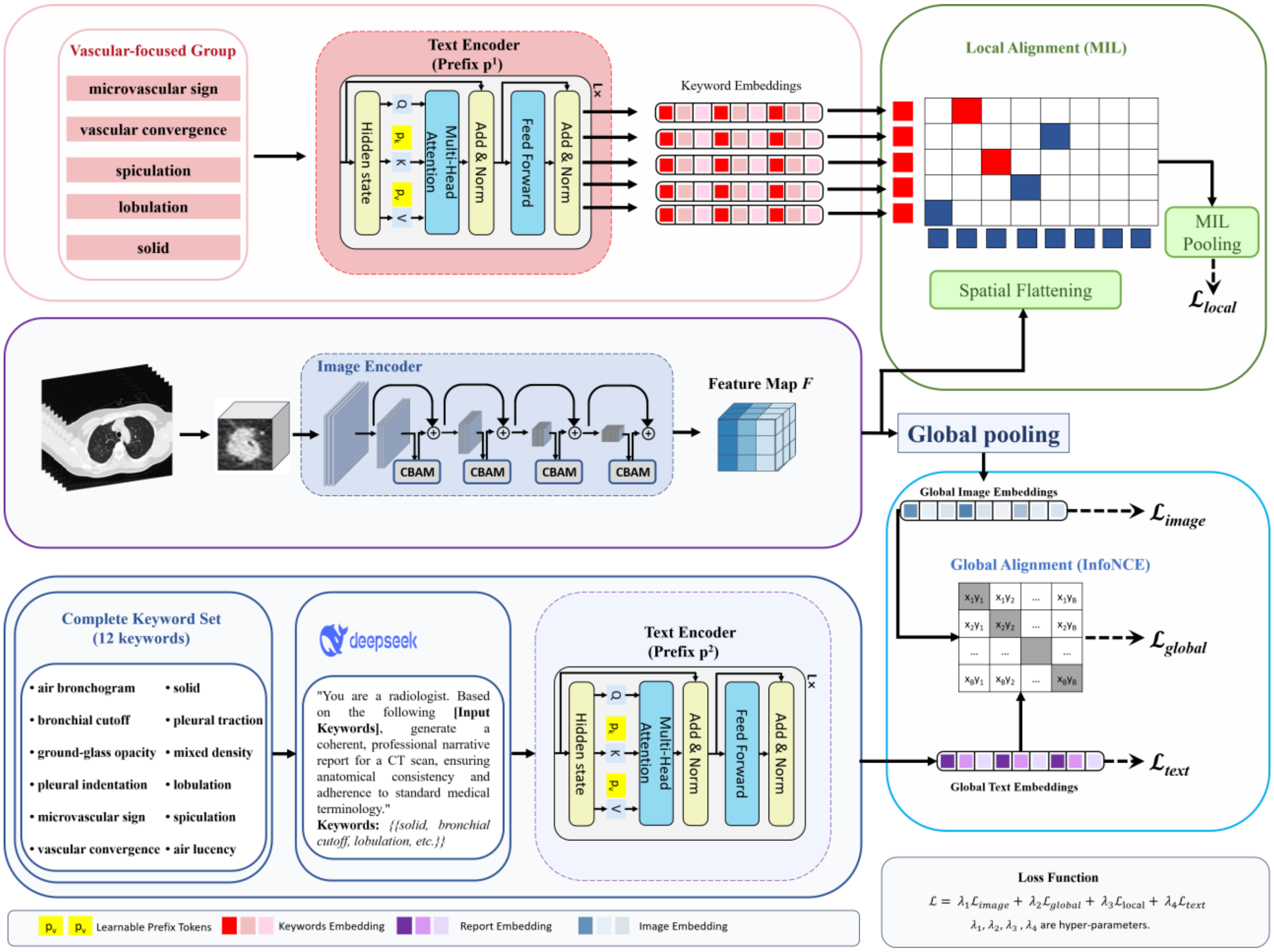

**Figure 2.** Overview of the proposed dual-granularity image–text alignment framework for preoperative prediction of lung adenocarcinoma invasiveness.

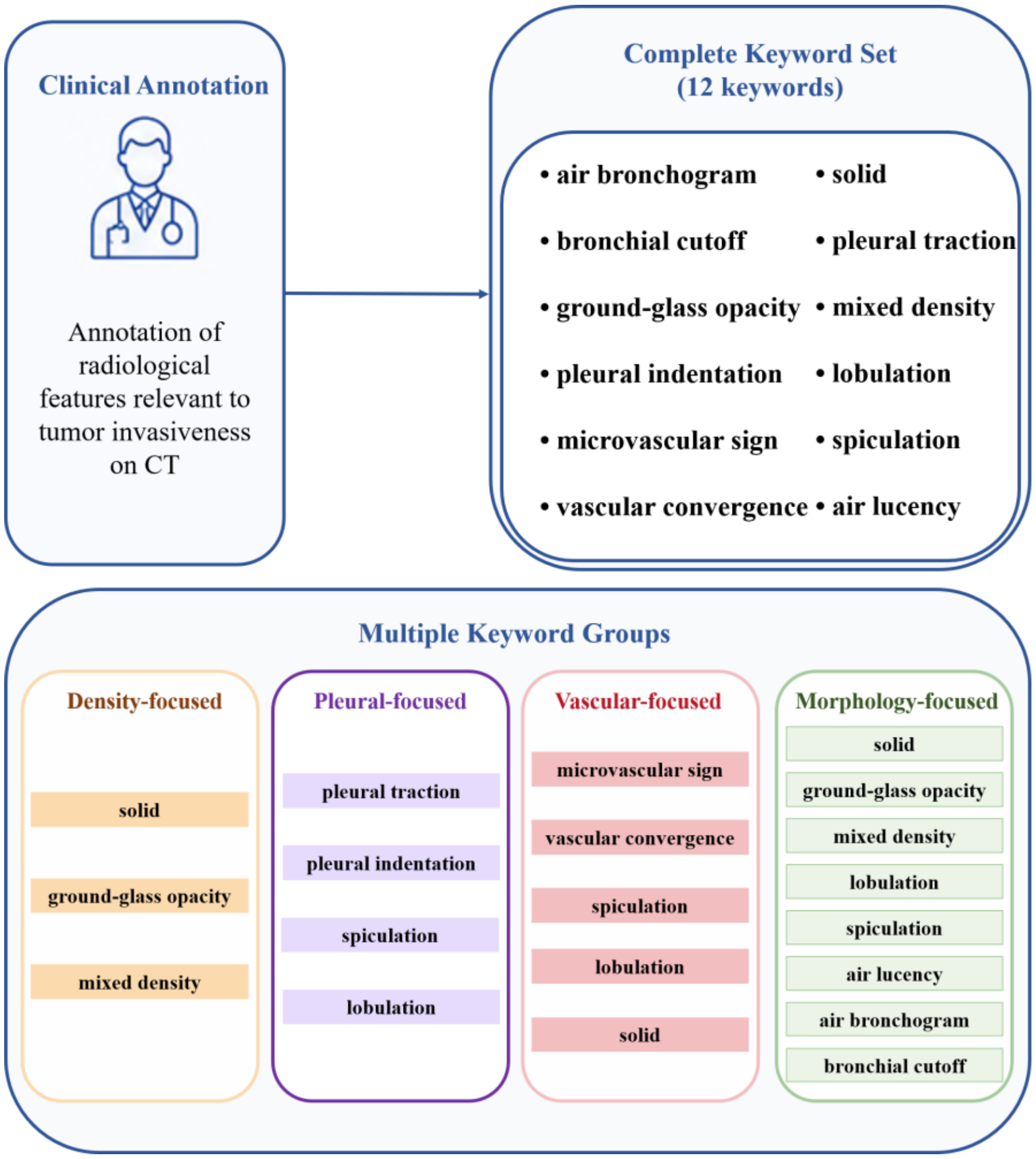

**Figure 3.** Overview of radiological keyword construction and grouping.

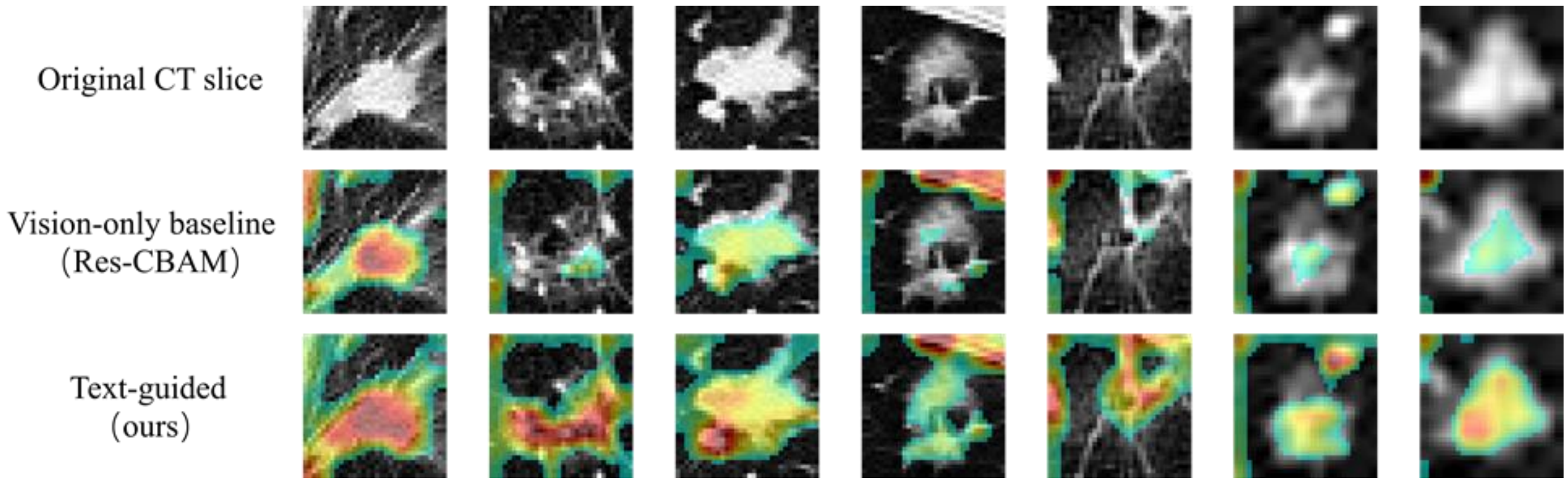

**Figure 4.** Representative Grad-CAM visualizations of the vision-only Res-CBAM baseline and the proposed text-guided framework. Each column represents a different pulmonary nodule. The top row shows the original CT images, and the middle and bottom rows show the activation maps for Res-CBAM and the proposed framework, respectively. Warmer colors indicate stronger model activation.